%% file: main.tex
\documentclass[letterpaper, 10 pt, journal, twoside]{ieeetran}
\usepackage{graphics} % for pdf, bitmapped graphics files
\usepackage{amsmath} % assumes amsmath package installed
\usepackage{amssymb}  % assumes amsmath package installed
\usepackage{booktabs}

\usepackage{multirow}
\usepackage{tabularx}
\usepackage{graphicx}
\usepackage{adjustbox}
\usepackage{subcaption}
\usepackage{xcolor}
\usepackage[font=footnotesize]{caption}
\usepackage{makecell}

\usepackage{pifont}
\usepackage{booktabs}  % for \toprule, \midrule, \bottomrule
\newcommand{\stdgray}[1]{\textcolor{gray}{\scriptsize(±#1)}}
\newcommand{\meanstd}[2]{#1\;\stdgray{#2}}
\newcommand{\bmeanstd}[2]{\textbf{#1}\;\stdgray{#2}}

\usepackage[final]{changes}
\definechangesauthor[name={Sehoon Ha}, color=blue]{SH}
\definechangesauthor[name={Morgan Byrd}, color=red]{MB}
\setdeletedmarkup{}   % deleted text is kept in the source but not rendered

\title{\LARGE \bf
CF-JEPA: Improving Robustness of JEPA World Models via Controllability Factorization
}

\author{Morgan Byrd$^{1*}$, Robert Wright$^{2}$, and Sehoon Ha$^{1}$% <-this % stops a space
\thanks{$^{1}$Georgia Institute of Technology, Atlanta, GA, 30308, USA}%
\thanks{$^{2}$Georgia Tech Research Institute, Atlanta, GA, 30308, USA}%
\thanks{*Correspondence to abyrd45@gatech.edu}
}

\begin{document}

\maketitle

%%%%%%%%%%%%%%%%%%%%%%%%%%%%%%%%%%%%%%%%%%%%%%%%%%%%%%%%%%%%%%%%%%%%%%%%%%%%%%%%
\begin{abstract}
Controlling an agent with vision requires being able to separate useful information from irrelevant background information. JEPA-style latent world models seem like a natural approach for this, as they do not perform pixel-level reconstruction; however, they are still sensitive to these distractor signals and experience latent collapse. In this work, we introduce Controllability Factorized JEPA (CF-JEPA), a JEPA-style world model which splits the latent space into controllable and uncontrollable subspaces. This factorization allows us to capture all the distractor information into the uncontrollable region, while we use the control-relevant latent information for our task. With this, we show comparable performance across 2D and 3D control tasks under nominal conditions and improved performance under distracted conditions, where CF-JEPA is the only model that does not experience latent collapse. We also validate our model under distracted conditions for a simulated robot task, highlighting the practical application of such a scheme.

\end{abstract}


\input{section/introduction}

\input{section/related-work}
\input{section/method}
\input{section/results}
\input{section/conclusion}

\vspace{-1.em}

\section{Acknowledgments}
Claude was used for generating code and running experiments. GPT was used for minor edits. All content was reviewed and validated by the authors.

\bibliographystyle{IEEEtran}
\bibliography{main.bib}

\end{document}

%% file: section/introduction.tex
\section{INTRODUCTION}
\label{S:1}

Humans are capable of performing tasks using visual control, regardless of what sort of background visual distractor information or noise might be present. Random objects moving into the visual field can be easily ignored and do not interfere \added[id=SH]{with} task progress or break our understanding of the world. \added[id=SH]{Robots deployed in diverse environments should likewise accomplish their tasks despite such visual distractors.} Thus, a reasonable expectation of a generative world model might be to have a similar invariant nature to these kinds of external visual stimuli.

% Teaser: the four tasks as a 2x2 grid, cropped from images/Tasks.pdf (960x540 bp).
\begin{figure}[t]
\centering
\begin{minipage}{0.9\columnwidth} % overall teaser width
\includegraphics[width=0.49\linewidth, trim={13 157 737 172}, clip]{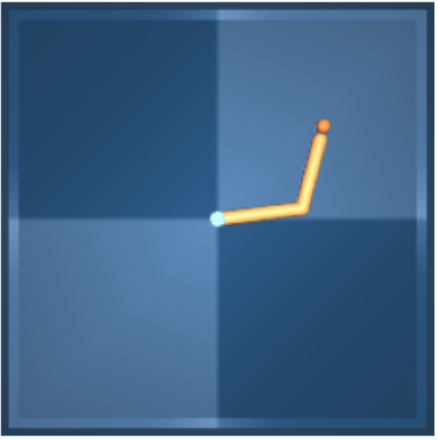}\hfill
\includegraphics[width=0.49\linewidth, trim={255 158 497 173}, clip]{images/Tasks.pdf}\\[0.5ex]
\includegraphics[width=0.49\linewidth, trim={495 157 255 172}, clip]{images/Tasks.pdf}\hfill
\includegraphics[width=0.49\linewidth, trim={736 157 14 172}, clip]{images/Tasks.pdf}
\end{minipage}
\caption{Primary tasks for evaluation. We test on the DM Control Suite Reacher, PushT, OGBench Cube, and TwoRoom tasks.}
\label{fig:tasks}
\vspace{-0.75em}
\end{figure} 

However, typical world models~\cite{Hafner2019, iris2023} struggle under these conditions as they rely on \added[id=SH]{pixel-level} reconstruction \added[id=SH]{to learn their representations}, and reconstructing such random content is difficult and wasteful. There have been attempts to handle this~\cite{okada2021, zhang2020learning, fu2021learning}, but they often still rely on reconstructing the scene along with additional \added[id=SH]{techniques} or rely on assumptions that may not hold in realistic robotic scenarios. 

\added[id=SH]{Joint-Embedding} Predictive Architectures (JEPA) \added[id=SH]{appear to be} a natural way to handle this kind of distraction, as they remove the generative model built on pixel-level reconstruction and predict dynamics \added[id=SH]{in latent space} instead. However, we find that such JEPA world models~\cite{leworldmodel2026} like LeWM are still very susceptible to task-irrelevant visual distractor \added[id=SH]{signals}, and the latent space collapses as a result. Incorporating an inverse dynamics loss as in Sensorimotor World Model (SMWM)~\cite{ivashkov2026sensorimotor} improves robustness, but it still uses the full latent rather than separating distractor elements from control.

Building \added[id=SH]{on} the JEPA \added[id=SH]{framework}, we propose CF-JEPA, a world model with additional latent structure for improved robustness under visual distractor signals. We incorporate controllability factorization, where we partition the latent space into a control-relevant, action-conditioned region and a control-irrelevant, action-free region. This explicit \added[id=SH]{partition} \added[id=SH]{has a minimal to moderate impact on} nominal performance and offers \added[id=SH]{a} strong benefit in the presence of visual distractors, where \added[id=SH]{CF-JEPA is the only tested model that} does not collapse.

\added[id=SH]{We evaluate CF-JEPA against LeWM and SMWM on four visual control tasks and a simulated robot task. With two visual distractors, LeWM collapses on all tasks, and SMWM collapses on Reacher, where its success rate drops to 0.10 while CF-JEPA retains 0.48. As the number of distractors grows, SMWM degrades rapidly, and its success rate on PushT and TwoRoom falls from 0.86 and 0.98 respectively to 0.27 and 0.46 as it collapses, while CF-JEPA maintains 0.57 and 0.82 respectively. We further hypothesize that the benefit of controllability factorization is most evident when the action space requires perceiving the state of the agent, and we confirm this on the simulated robot task. Finally, a diagnostic decoder on Reacher shows that the controllable latent of CF-JEPA preserves the agent, whereas the baseline latents lose the agent.}

Our key contributions are the following:

\begin{enumerate}
    \item CF-JEPA, a simple world model architecture that improves robustness to irrelevant visual distractors via controllability factorization, where we separate the irrelevant, uncontrollable environment features from the relevant, controllable features in our latent space.

    \item We thoroughly compare against benchmarks and show that we match under nominal conditions and outperform under visual perturbations, across a suite of 2D and 3D control tasks, along with a simulated robot task.

    \item We analyze the latent space to \added[id=SH]{examine the effect} of the controllability factorization.

    % \item We also benchmark the performance of all models to bound scope of benefits from model improvements versus control improvements on harder robotics tasks from ManiSkill~\cite{taomaniskill3}.

    % \morgan{I think given space requirements and lack of overall success, we can remove this point.}
    % \sehoon{Please write this one more clearly. It is hard to tell what the claim is.}
\end{enumerate}

%% file: section/related-work.tex
\section{RELATED WORK}
\label{Relatedwork}

\subsection{World Models}
World models~\cite{ha2018worldmodels} are a way of learning the dynamics of an environment via generative models directly from data. Learning the latent representation via reconstruction of the input pixels is a common method, seen in the Dreamer~\cite{Hafner2019, Hafner2020, Hafner2023} series of models, along with many others including \cite{iris2023, fu2021learning, Byrd2025, lee2020slac, robine2023transformerbased}. Other methods for learning models without reconstruction include contrastive learning~\cite{okada2021, okada2022} and prototypical representations~\cite{pmlr-v162-deng22a}. More recently, joint-embedding predictive architectures (JEPA)~\cite{assran2023selfsupervisedlearningimagesjointembedding} have gained prominence by constructing the model via predicting feature-space targets rather than pixels. JEPA style models vary widely and include many methods for preventing collapse, including using a pretrained encoder as in Dino-WM~\cite{zhou2024dinowmworldmodelspretrained}, via a VICReg loss~\cite{bardes2022vicregvarianceinvariancecovarianceregularizationselfsupervised} as in PLDM~\cite{sobal2025learning}, and via a SIGReg~\cite{balestriero2025lejepaprovablescalableselfsupervised} loss as in LeWM~\cite{leworldmodel2026}. We build off LeWM for our work, where we add an inverse dynamics loss and augment the latent space with additional structure in order to improve robustness.

\subsection{Inverse Dynamics for World Models}
Inverse dynamics is the process of computing the required forces or torques to produce a given reference trajectory. It is popularly used in robotics for control of manipulator arms~\cite{mahmoodabadi2019inverse, rodriguez1987kalman} and walking robots~\cite{zapolsky2014, righetti2011, mistry2010} as well as used in guidance application~\cite{ghosh2000nonlinear}. In machine learning, it is leveraged for computing actions given a reference video~\cite{torabi2018behavioralcloningobservation, pathak2018zeroshotvisualimitation} or motion capture sequence~\cite{liu2025imdyhumaninversedynamics} and for representation learning~\cite{Lesort_2018, young2025, brandfonbrener2023inversedynamicspretraininglearns}. For world modeling specifically, learning an inverse dynamics model is commonly used for acting in imagination, as in~\cite{Hafner2019, sutton1991}, as well as representation learning, as in~\cite{ivashkov2026sensorimotor, zhu2025uwm, acmtm2026}. The most relevant case for our work is the concurrent work Sensorimotor World Models (SMWM)~\cite{ivashkov2026sensorimotor}, which builds off LeWM, removing the SIGReg loss and adding an inverse model as an anti-collapse mechanism. We also leverage an inverse dynamics loss, but we use it to structure the latent space into controllable and uncontrollable subspaces, giving improved results in terms of robustness over the standard inverse dynamics loss.

\subsection{Distractor Robustness}
Rejecting irrelevant distractor signals and focusing on only task relevant information is key for effective visual control. The reconstruction free world model methods above tackle this by not spending capacity modeling every pixel but capturing the underlying state~\cite{okada2021, pmlr-v162-deng22a}. In model-free RL, bisimulation~\cite{zhang2020learning, ferns2012metricsfinitemarkovdecision} is one method for generating useful, robust representations from pixel based environments without reconstruction. Task Informed Abstractions~\cite{fu2021learning} and Denoised MDPs~\cite{wang2022denoised} partition the latent space similarly to ours, but still require reconstruction and thus still expend some model capacity for representing the pixel level view of the environment, which can be especially difficult if there are distractor signals. We avoid pixel level reconstruction and gain robustness via separating the controllable latent subspace from the uncontrollable portion, and using only the controllable section for control.

\vspace{-1em}

%% file: section/method.tex
\section{METHOD}

\subsection{Problem Definition}
\label{ssec:problem}
% \added[id=SH]{We study control from pixels without rewards. For each task, we assume an offline dataset $\mathcal{D}=\{(x_t, a_t, x_{t+1})\}$ of transitions, where $x_t$ is an RGB image and $a_t$ is a continuous action. The trajectories carry no reward and no task label, and we place no optimality requirement on the policy that produced them. They only need to cover the dynamics of the environment. From this data, we learn an encoder $f$ that maps an image to a latent vector and a dynamics model that predicts the next latent from the current one and the action. At test time, the agent receives a goal image $x_g$, plans in the latent space, and we report the fraction of trials in which it reaches the goal.}

We study control from pixels without reward data. For each task, we assume access to an offline dataset $\mathcal{D}=\{(x_t, a_t, x_{t+1})\}$ of transitions, where $x_t$ is an RGB image and $a_t$ is the continuous action that maps from $x_t$ to $x_{t+1}$. The trajectories carry no reward information or task label, and we do not require an optimal policy for generation. Sufficient coverage for accurate representation of the environment dynamics is the primary requirement. From this data, we learn an encoder $f$ that maps the input image $x_t$ to a latent vector $z_t$ as well as a dynamics model that predicts the next latent $z_{t+1}$ using the current latent $z_t$ and action $a_t$. At inference, the agent receives a goal image $x_{goal}$, plans in the latent space using model-predictive control via our world model, and we register the success rate as the percentage of trials that successfully get within a threshold of the goal.

% \added[id=SH]{We assume that the environment has a factorized structure. The underlying state splits into a controllable part $s^c$ and an uncontrollable part $s^u$. The action moves only the controllable part, the two parts evolve independently of each other, and the agent sees neither of them directly. What it sees is an image rendered from both:}

We prescribe a factorized structure for the environment MDP dynamics. We split the underlying state $s$ into a controllable part $s^c$ and an uncontrollable part $s^u$. The action input influences only the controllable section, with the uncontrollable section dynamics progressing autonomously, entirely separate from the controllable dynamics. The agent has no direct access to either state, and only sees the image rendered from them:

% \begin{addedeq}
\begin{equation}
\begin{aligned}
    T(s' \mid s, a) &= T_c(s^{c\prime} \mid s^c, a)\; T_u(s^{u\prime} \mid s^u), \\
    x_t &= q(s^c_t, s^u_t).
\end{aligned}
\end{equation}
% \end{addedeq}
% \added[id=SH]{Here $q$ is the emission function that renders a state into an image. A factorized state paired with such a joint emission is called an exogenous block MDP~\cite{efroni2022exbmdp}. We further assume that visual distractors live in $s^u$. A distractor enters the image through $q$, it follows its own dynamics $T_u$, and no action of the agent can change it. A shape drifting across the background is a distractor, while the arm the agent moves is not.}

Here $q$ is the emission function that renders a state $s$ into an image $x$. This structure of a factorized state paired with a joint emission is called an exogenous block MDP~\cite{efroni2022exbmdp}. \added[id=MB]{Note, this is a strict generalization of the standard MDP format and, if $T_u = I$, $s^u=\emptyset$, and $s^c = s$, would reduce to a standard MDP.} 

We further assume that visual distractors are captured in the uncontrollable $s^u$. A distractor signal enters the image through $q$, it follows its own dynamics $T_u$, and no action of the agent can change it. \added[id=SH]{Movement in the image background is a distractor, while the arm controlled by the agent is not.}

% \added[id=SH]{Our goal is a world model that is robust against variation in $s^u$, so that control performance holds up as the uncontrollable part of the scene changes. This factorization is an idealization. It holds when a distractor only adds pixels to the image, and it breaks down when the distractor interferes with the task itself, for example by occluding it.}

Our goal is to learn a world model that is robust against variation in $s^u$, such that we maintain high quality control performance even when the uncontrollable state changes. Under ideal circumstances, the factorization perfectly captures the control relevant information and rejects all other information. In reality, it works well for things like extraneous background pixels or noise, but would be incapable of capturing things that directly interfere with the task, like occlusion of the task itself.

\begin{figure*}[t]
\begin{center}
\includegraphics[width=0.8\linewidth, trim={0.25cm 4.75cm 0.75cm 0.cm}]{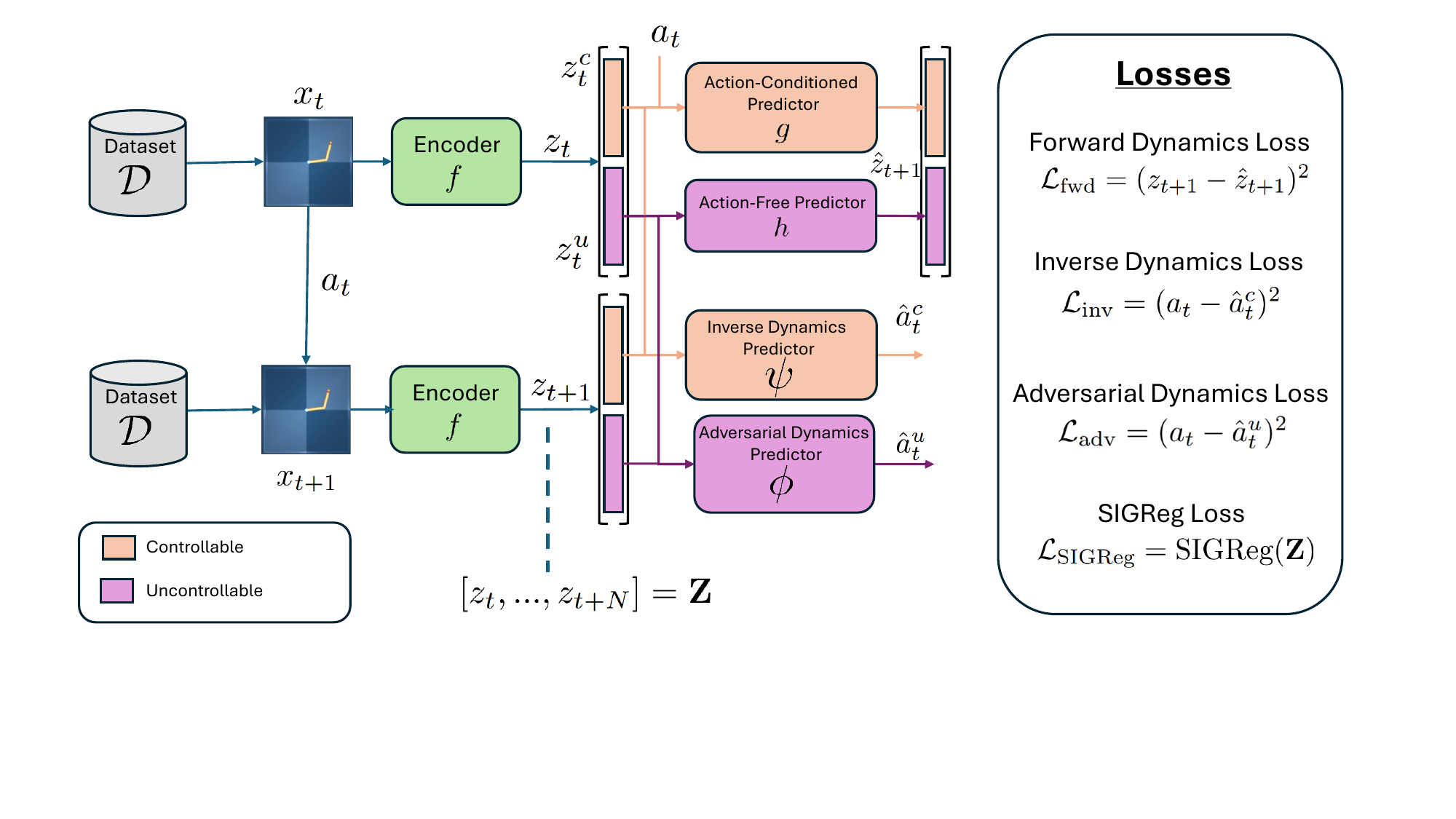}
\end{center}
\caption{Overview of our method.}
\label{fig:overview}
\vspace{-1em}
\end{figure*}

\subsection{Background}
\added[id=SH]{LeWM~\cite{leworldmodel2026} learns a latent world model end to end from pixels. It has two parts, an encoder and a predictor,}
% \begin{addedeq}
\begin{equation}
    z_t = \text{enc}_\theta(x_t), \qquad \hat{z}_{t+1} = \text{pred}_\varphi(z_t, a_t),
\end{equation}
% \end{addedeq}
% \added[id=SH]{and it trains them so that the predicted latent matches the encoded next frame. This objective alone has a trivial solution. An encoder that maps every image to the same vector drives the error to zero and keeps no information about the observation. This failure is called representation collapse. LeWM rules it out with SIGReg~\cite{balestriero2025lejepaprovablescalableselfsupervised}, which pushes the distribution of latents toward an isotropic Gaussian along random one-dimensional projections.}

and it trains them so that the predicted latent $\hat{z}_{t+1}$ matches the encoded next frame $\text{enc}_\theta(x_{t+1})=z_{t+1}$. This objective $\mathcal{L} = (z_{t+1}-\hat{z}_{t+1})^2$ alone has a trivial solution. The encoder can simply map every image input to a constant vector output, making the loss zero but removing any information about the environment and providing no benefit for dynamics. This failure is an example of the representation collapse that comes from the JEPA objective without the full pixel reconstruction loss. LeWM bypasses this using SIGReg~\cite{balestriero2025lejepaprovablescalableselfsupervised}, which pushes the distribution of latents toward an isotropic Gaussian along random one-dimensional projections.

% \added[id=SH]{SMWM~\cite{ivashkov2026sensorimotor} removes SIGReg and regularizes with inverse dynamics instead. A small network predicts the action $a_t$ from the pair of latents $(z_t, z_{t+1})$, and it reads the entire latent vector. A collapsed encoder would hand this network the same input for every transition, so an accurate action prediction rules collapse out. The loss also keeps the latent action-informative.}

SMWM~\cite{ivashkov2026sensorimotor} removes SIGReg and regularizes with inverse dynamics to prevent collapse instead. A small MLP is given a pair of consecutive latents $[z_t, z_{t+1}]$ and predicts the action $a_t$ that moved from $z_t$ to $z_{t+1}$. Reducing this loss $\mathcal{L} = (\hat{a}_t-a_t)^2$ requires the latent space to not collapse, and it makes the latent space have action-relevant information while prescribing no distribution shape like SIGReg would.

% \added[id=SH]{Neither model structures the latent along the factorization of Sec.~\ref{ssec:problem}. Both encode the controllable and uncontrollable parts into a single vector, so a distractor enters the representation on the same footing as the agent.}

Neither model structures the latent along the factorization of Sec.~\ref{ssec:problem}. Both treat controllable and uncontrollable information equally, allowing for distractor signals to impact the representation and collapse the latent space.

\vspace{-0.5em}

\subsection{CF-JEPA: Learning Controllability Factorized World Models}
Where LeWM works well under nominal conditions across the tested tasks, it fails in the presence of visual disturbances. To address this, we modify the latent space formulation for increased robustness. Our insight is that we only need a certain amount of the information in the image in order to solve the control task, so we should arrange the learning objective with this as a goal. Our CF-JEPA explicitly factorizes the latent space for that goal, allowing us to use the control relevant latent subspace and ignore the other unnecessary visual information in a control irrelevant latent subspace.

Learning our world model proceeds similarly as in LeWM~\cite{leworldmodel2026}. We assume access to an offline dataset $\mathcal{D}$ of state-action transitions on a task-by-task basis. Given this reward free data, we then train an encoder $f$ which maps a given observation $x_t$ to a latent vector $z_t$. Once we have the latent vector, we train a separate predictor network which predicts the next latent vector given the current one and current action.

We structure our latent space into two regions, $z_t =$[$z_{t}^{c}, z_{t}^{u}$], where $z_{t}^{c}$ is a controllable region which should affect only the controllable part of the world and $z_{t}^{u}$ is an uncontrollable region which should handle everything else and explicitly not model the control. 

The learning objective of our model is as follows. 

\noindent\textbf{Forward Dynamics.} First, we have the forward prediction loss mentioned above. We compute this forward loss on a per-subspace basis, where the controllable region only influences the controllable region, and likewise for the uncontrollable region. Also, only the controllable region uses the action, as the uncontrollable region by construction is not influenced by the action. Specifically, we have one predictor $g$ that is action conditioned for $z_{t}^{c}$ and another predictor $h$ with no action conditioning for $z_{t}^{u}$. $g(a_t, z_{t}^{c}) = \hat{z}_{t+1}^{c}$, $h(z_{t}^{u}) = \hat{z}_{t+1}^{u}$\added[id=SH]{. The two predictions are concatenated back into $\hat{z}_{t+1} = [\hat{z}^{c}_{t+1}, \hat{z}^{u}_{t+1}]$.} and the loss is:

\begin{equation}
    \mathcal{L}_{\text{fwd}} = (z_{{t+1}} - \hat{z}_{{{t+1}}})^2
\end{equation}

\begin{figure*}[t]
\begin{center}
\includegraphics[width=0.8\linewidth, trim={0.cm 5.9cm 0.cm 0.25cm}]{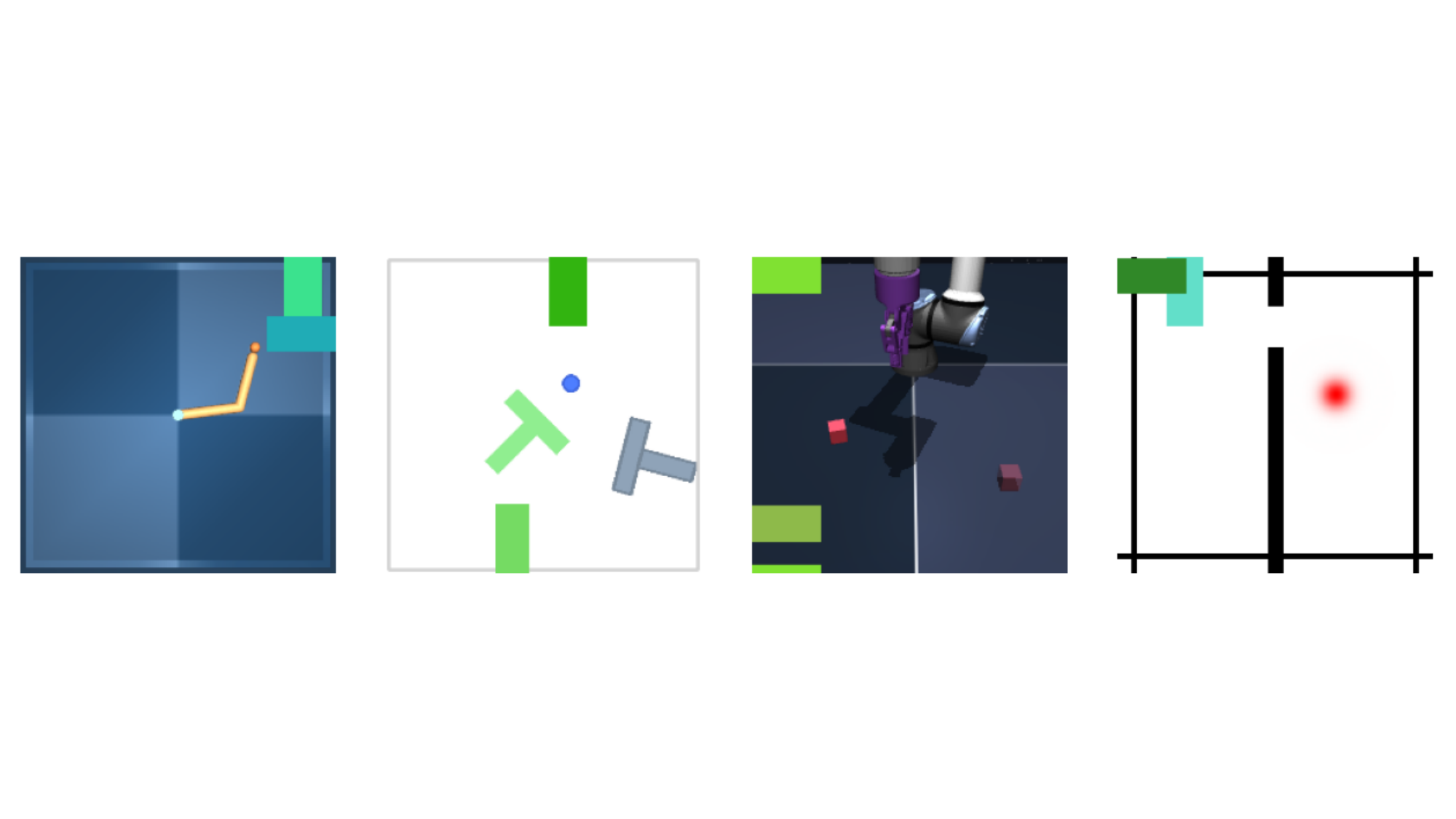}
\end{center}
\caption{Tasks with example distractors (N=2).}
\label{fig:tasks-with-distractor}
\vspace{-1em}
\end{figure*}

\noindent\textbf{Inverse Dynamics.} Next, we have an inverse dynamics loss. This is to set up the latent space controllability structure above and also to focus on the important aspect of the model, which is the ability to control downstream tasks. Specifically, we use another MLP predictor $\psi$ that predicts the action given the current and next controllable latent, \added[id=SH]{$\psi(z_{t}^{c}, z_{t+1}^{c}) = \hat{a}^{c}_t$}.

% \begin{addedeq}
\begin{equation}
    \mathcal{L}_{\text{inv}}=(a_t-\hat{a}^{c}_{t})^2
\end{equation}
% \end{addedeq}

\noindent\textbf{Adversarial Dynamics.} Then, we have an adversarial dynamics loss. This is to bolster the latent space structure by penalizing action predictability of the uncontrollable space. Here, we proceed similarly as with the inverse dynamics loss, using a MLP predictor $\phi$ to predict actions between the uncontrollable latents, but \added[id=SH]{prepend a gradient reversal layer~\cite{ganin2015unsuperviseddomainadaptationbackpropagation} to $\phi$} to actively penalize effective prediction, \added[id=SH]{$\phi(z_{t}^{u}, z_{t+1}^{u}) = \hat{a}^{u}_t$}.
% \begin{addedeq}
\begin{equation}
    \mathcal{L}_{\text{adv}}=(a_t-\hat{a}^{u}_{t})^2
\end{equation}
% \end{addedeq}

\noindent\textbf{SIGReg.} Finally, we utilize the same SIGReg loss as in LeWM for preventing collapse of the latent space. A schematic can be seen in Fig.~\ref{fig:overview}. The full objective is the following:

\begin{equation}
    \mathcal{L} = \mathcal{L}_{\text{fwd}} + \alpha \mathcal{L}_{\text{inv}} + \beta \mathcal{L}_{\text{adv}} + \gamma \mathcal{L}_{\text{SIGReg}}
\end{equation}
where for all tasks $\beta=1$ and $\gamma=0.09$, while $\alpha$ was specified on a task by task basis. \added[id=SH]{The lower block of Table~\ref{table:models} lists these per-task weights, alongside the weight $\lambda$ that SMWM tunes for its own inverse dynamics loss.}

% \added[id=SH]{Table~\ref{table:models} also summarizes how the three models differ. LeWM relies on the forward loss and SIGReg alone, SMWM replaces SIGReg with an inverse dynamics loss over the full latent, and we keep SIGReg while restricting the inverse dynamics loss to $z^c$ and adding the adversarial loss on $z^u$.}

Table~\ref{table:models} also summarizes how the three models differ. LeWM relies on the forward loss and SIGReg alone, SMWM replaces SIGReg with an inverse dynamics loss over the full latent, and we keep SIGReg while restricting the inverse dynamics loss to $z^c$ and adding the adversarial loss on $z^u$. \added[id=MB]{Also, since we factorize the control-relevant components explicitly, we use only $z^c$ for control, while both others use the full latent $z$.}

\input{tables/inverse-weights}

% \begin{figure*}[t]
% \begin{center}
% \includegraphics[width=\linewidth, trim={0.25cm 4.75cm 0.75cm 0.cm}]{images/Overview.pdf}
% \end{center}
% \caption{Overview of our method. \protect\sehoon{Please check the font size.} \morgan{Need to update to match notation with rest of paper. Can probably shrink for space purposes.}}
% \label{fig:overview}
% \vspace{-1em}
% \end{figure*}

% \sehoon{This last part could be its own named paragraph, or even a subsection.}
\noindent\textbf{Control.} For control, we plan using the world model in latent space. We optimize via model-predictive control using the cross entropy method~\cite{RUBINSTEIN_CEM}. Given a goal image, we compute the goal latent $z_{goal}=f(x_{goal})$ and compute the cost based on our distance in latent space from that goal, accumulating the cost over the predicted latent $z_k^c$ at each step $k$ in the receding horizon control.

\begin{equation}
    C_{CEM} = \sum_{k=0}^K\|z_k^c-z_{goal}^c\|
\end{equation}

\subsection{Implementation Details}
Our model architecture follows heavily from LeWM~\cite{leworldmodel2026}.

% \begin{figure*}[t]
% \begin{center}
% \includegraphics[width=0.8\linewidth, trim={0.cm 5.7cm 0.cm 5.25cm},clip]{images/Tasks.pdf}
% \end{center}
% \caption{Primary tasks for evaluation. We test on the DM Control Suite Reacher, PushT, OGBench Cube, and TwoRoom tasks.}
% \label{fig:tasks}
% \vspace{-1em}
% \end{figure*}

% \begin{figure*}[t]
% \begin{center}
% \includegraphics[width=0.8\linewidth, trim={0.cm 5.9cm 0.cm 0.25cm}]{images/Tasks-with-Distractors.pdf}
% \end{center}
% \caption{Tasks with example distractors (N=2).}
% \label{fig:tasks-with-distractor}
% \vspace{-1em}
% \end{figure*}

\noindent\textbf{Encoder.} The encoder \added[id=SH]{$f$} is a vision transformer, using the ViT-Tiny model from HuggingFace, with a patch size of 14. The output latent vector $z$ is 192 dimensional, split so $z^c\in\mathbb{R}^{128}$ and $z^u\in\mathbb{R}^{64}$.

\noindent\textbf{Forward model.} The forward model consists of two transformer predictor networks, one action-conditioned for the controllable latent \added[id=SH]{($g$)} and one action-free for the uncontrollable latent \added[id=SH]{($h$)}. The controllable network uses learned position embeddings and causal attention along with AdaLN-zero conditioning on the action to predict $z_{t+1}^c$ given the input $[z_t^c, a_t]$. The action-free network uses a smaller transformer with causal attention to predict $z_{t+1}^u$ given the input $z_t^u$.

\noindent\textbf{Inverse model.} The inverse model \added[id=SH]{$\psi$} consists of a two layer MLP network with GELU activations that maps $[z_t^c, z_{t+1}^c]$ to $a_t$.

\noindent\textbf{Adversarial model.} The adversarial model \added[id=SH]{$\phi$} consists of a two layer MLP network with GELU activations augmented with a gradient reversal layer that maps $[z_t^u, z_{t+1}^u]$ to $a_t$.

\input{tables/nominal-task}

\noindent\textbf{Training.} We train all models using AdamW with warmup using a learning rate of $5e-5$ and weight decay of $1e-3$. We train for 10 epochs on each task using a batch size of 128. All training was done using a machine with a single NVIDIA 5090 GPU and each model took less than 8 hours to train. 

\noindent\textbf{Planning.} For CEM planning, we consider 300 candidate actions per iteration and 30 iterations per planning session. We plan out five actions and replan over a receding horizon of five.

\added[id=MB]{\noindent\textbf{Decoder. (Diagnostic Only)} For latent space analysis after training, we also trained a decoder to map the latent vector $z$ from the trained frozen encoder $f$ to the pixel inputs $x$. The decoder is a stack of transposed convolutions that maps to the input image, in a manner similar to DCGAN~\cite{radford2016unsupervisedrepresentationlearningdeep}.}

% \vspace{-0.5em}

%% file: tables/inverse-weights.tex
\begin{table}[t]
\centering
\renewcommand{\arraystretch}{1.15}
\caption{\protect\added[id=SH]{Model components and the loss weight that is tuned per task, $\lambda$ for the inverse dynamics loss of SMWM and $\alpha$ for our inverse loss. LeWM has neither, so no weight applies.}}
\label{table:models}
\begin{tabular}{@{}lccc@{}}
\toprule
 & \textbf{LeWM} & \textbf{SMWM} & \textbf{CF-JEPA} \\
\midrule
\multicolumn{4}{@{}l}{\textit{Model components}} \\
\addlinespace[1pt]
Forward loss          & O & O & O \\
SIGReg                & O & X & O \\
Inverse dynamics      & X & O\,(full $z$) & O\,($z^c$) \\
Adversarial dynamics  & X & X & O\,($z^u$) \\
Latent factorization  & X & X & O \\
Control               & $z$ & $z$ & $z^c$ \\
\midrule
\multicolumn{4}{@{}l}{\textit{Per-task loss weight}} \\
\addlinespace[1pt]
Reacher      & -- & 5   & 10\\
PushT        & -- & 30  & 3\\
OGBench Cube & -- & 1   & 10\\
TwoRoom      & -- & 0.1 & 3\\
\bottomrule
\end{tabular}
\vspace{-0.5em}
\end{table}

%% file: tables/nominal-task.tex
\begin{table}[]
\vspace{0.3em}
\centering
\caption{Nominal task performance. We report the values for LeWM and SMWM directly from the source papers and list our values as mean $\pm$ standard deviation across three training seeds.}
\label{table:nominal-task-performance}
\begin{tabular}{l|l|l|l}
\hline
Task & LeWM & SMWM & CF-JEPA \\ \hline
Reacher & 0.86 & 0.66 & \meanstd{0.61}{0.03} \\ \hline
PushT & 0.96 & 0.83 & \meanstd{0.78}{0.04} \\ \hline
OGBench Cube & 0.74 & 0.84 & \meanstd{0.81}{0.03} \\ \hline
TwoRoom & 0.87 & 0.99 & \meanstd{0.95}{0.04} \\ \hline
\end{tabular}
\vspace{-0.5em}
\end{table}

%% file: section/results.tex
\section{Experimental Results}

% We evaluated our method on a set of tasks in order to evaluate the following research questions:
\added[id=SH]{We evaluate our method to answer the following research questions:}

\begin{enumerate}
    \item How does CF-JEPA perform against the baseline methods under nominal conditions?

    % \item How does incorporating the explicit factorization of controllability improve the control success under visually perturbed conditions?
    \item \added[id=SH]{How does controllability factorization improve control performance under visual distractors?}

    % \item How does this impact the latent space?
    \item \added[id=SH]{Does the robustness of CF-JEPA and SMWM hold when the action space requires perceiving the state of the agent?}

    \item \added[id=SH]{How does controllability factorization separate control-relevant information from distractors in the latent space?}

    % \item What are the limits of this visual representation learning for harder robotics control tasks?
\end{enumerate}

\vspace{-0.5em}

\subsection{Experimental Setup}

% We evaluate our method against two primary baselines. LeWM is a strong world model that we used for development which uses just the forward dynamics loss and SIGReg. We also compare against Sensorimotor World Model (SMWM), which has a similar inverse dynamics loss as our method, without any factorization, and removes the SIGReg loss, relying on the inverse dynamics alone for regularization and preventing latent collapse. 
\added[id=SH]{We compare CF-JEPA against two baselines. LeWM~\cite{leworldmodel2026} is the base model that our method builds on, and it shows strong planning performance across our tasks. SMWM~\cite{ivashkov2026sensorimotor}, in contrast, makes its latent action-informative through an inverse dynamics loss, so we expect it to be more robust to visual disturbances \added[id=MB]{which do not impact the action}. Comparing against both lets us test whether controllability factorization adds robustness beyond these state-of-the-art models.}

% For the main four tasks in Figure~\ref{fig:tasks}, we use the datasets provided by \cite{leworldmodel2026}. The models were trained and evaluated using code adapted the base code of \cite{maes_lld2026swm} and \cite{balestriero2025stable}.

\added[id=SH]{For the four tasks in Fig.~\ref{fig:tasks}, we use the datasets released with LeWM~\cite{leworldmodel2026}. Each dataset was collected with an established policy for its task, ranging from expert demonstrations to scripted and learned controllers. Our method is agnostic to how the data is collected, and it only requires that the trajectories cover the dynamics of the environment (Sec.~\ref{ssec:problem}). We train and evaluate all models with code adapted from~\cite{maes_lld2026swm, balestriero2025stable}.}

The datasets are as follows:
\begin{enumerate}
    \item PushT, from~\cite{zhou2024dinowmworldmodelspretrained}. We use 2,300,000 frames across 18,000 episodes collected from expert demonstrations.

    \item OGBench Cube, from~\cite{ogbench_park2025}. We use 2,000,000 frames across 10,000 episodes collected \added[id=SH]{with a scripted policy}.

    \item TwoRoom, from~\cite{sobal2025stresstesting}. We use 920,000 frames across 10,000 episodes of noisy scripted data.

    \item Reacher, from~\cite{tassa2018deepmindcontrolsuite}. We use 2,000,000 frames across 10,000 episodes collected using SAC~\cite{haarnoja2018softactorcriticoffpolicymaximum}.
\end{enumerate}

% \morgan{Plus some extra ManiSkill stuff if useful. This will not just match the exact LeWM results.}

\vspace{-0.5em}

\subsection{\protect\added[id=SH]{Performance under Nominal Conditions}}
\label{ssec:nominal}
% Task performance across all baselines and all tasks under nominal conditions is shown in Table~\ref{table:nominal-task-performance}. In general, we closely match the performance of both baselines with some fluctuation on a task-by-task basis, although it is slightly lower, likely due to the effort of factorizing.
\added[id=SH]{Under nominal conditions (Table~\ref{table:nominal-task-performance}), CF-JEPA ranks second on OGBench Cube and TwoRoom, within 0.04 of the best model, and third on Reacher and PushT. On the latter two tasks, it trails LeWM by 0.25 and 0.18, respectively.} \added[id=MB]{Since we track very closely to SMWM, we} \added[id=SH]{hypothesize that this gap arises from the inverse dynamics objective shared by SMWM and CF-JEPA, which requires the latent to encode the dynamics of the agent in addition to predicting future states. This performance degradation is therefore expected, and the ranking changes substantially once visual distractors are introduced, as shown in Sec.~\ref{ssec:distractor}.}

\subsection{Performance under Visual Perturbation}
\label{ssec:distractor}

\subsubsection{Distractor Configuration}
To evaluate how our method captures the underlying controllable dynamics and ignores irrelevant features, we consider the case of perturbing the scene with additional visual distractors. Specifically, we add rectangles with a single randomly chosen color which are placed outward from a randomly chosen edge of the image frame. Each distractor is 49 pixels long perpendicular to the edge of the \added[id=SH]{$224\times224$} image and the width is a random value chosen in the range $[18,36]$ pixels. \deleted[id=SH]{The length is chosen so that the rectangle is a distracting element while not encroaching too much into the frame such that the task is occluded.} By default, two rectangles are used, although that number N is a hyperparameter that we adjust later. Each rectangle moves along the image edge with a constant random velocity in the range $[3, 18]$ pixels/step. The affected area of the frame is chosen such that the task is not occluded, so under ideal conditions, control performance should be unaffected. An example of each task with the distractor is shown in Fig.~\ref{fig:tasks-with-distractor}.

\input{tables/task-with-distractors}

\subsubsection{Standard Distractor Performance}
\added[id=SH]{To demonstrate the robustness of CF-JEPA to visual distractors, we evaluate all models on the four tasks under the standard distractor setting with $N=2$.} Table \ref{table:task-with-distractor-performance} shows the results across the different tasks for our method and the baselines. All models were trained with images augmented with N=2 distractors. The data is collected over three seeds. \added[id=SH]{The baseline LeWM degrades on every task, and it collapses on Reacher and PushT, where its success rate falls below 0.1.} SMWM performs much stronger, as the inverse dynamics loss also provides an amount of robustness, but also collapses under distractors in the Reacher task. Only our model did not collapse for any task, showing the benefit of the factorization for improving robustness against task-irrelevant distractors. \added[id=SH]{CF-JEPA even performs better with distractors than under nominal conditions on PushT (0.87 vs.~0.78) and OGBench Cube (0.85 vs.~0.81).} \added[id=MB]{The OGBench Cube results can likely be ascribed to the variance from the training seeds, but the PushT results are larger than that. Our hypothesis for this gap is that SIGReg forces both $z^c$ and $z^u$ to be populated and PushT under nominal conditions has no uncontrollable information, so some important information gets put into $z^u$, negatively impacting control. Under distractor conditions, this is not a problem as the distractor information can populate $z^u$. This is also not a problem for the other tasks as they include additional, uncontrollable background information that can get put into $z^u$.}

\added[id=MB]{For some quantitative support to this theory, we probed $z^c$ for both the nominal and distractor trained models to see how well it encoded the proprioceptive state. We see a $0.07 \pm 0.03$ increase in $R^2$ for decoding the block angle and a $0.1 \pm 0.02$ increase in $R^2$ for decoding the block position when going from the nominal to distractor case. This difference plausibly accounts for the control improvement under distractor conditions. Another important feature is that this distractor trained model can be deployed in the nominal environment and we still see better results, with a success rate of 0.87 $\pm$ 0.02.}

% \sehoon{Very interesting, but mostly conjecture. I would simplify this argument, while toning down some strong statements (e.g., toy setting...), unless we have additional data.}

% \input{tables/task-with-distractors}

\subsubsection{Extreme Distractor Performance}
Since Reacher was the only task where SMWM failed, we did further examination to determine the cause. Our hypothesis is that Reacher is a more difficult task for a vision-based agent to complete as the arm is thin and takes up little of the screen, so there is little for the model to leverage for dynamics learning and little to help prevent collapse under the addition of visual perturbations. To further analyze this, we trained additional models with increased distractor number N in order to determine if the distractor tolerance was task-specific and Reacher was a particularly low robustness task. \added[id=SH]{Table~\ref{table:distractor-sweep}} shows the performance per model across increased distractor values across tasks.

\input{tables/task-with-distractor-sweep}

% Figures \ref{fig:reacher-distractor-collapse}, \ref{fig:pusht-distractor-collapse}, \ref{fig:ogb-distractor-collapse}, and \ref{fig:tworoom-distractor-collapse} shows how the control performance of each task collapses as a function of the distractor number N. \morgan{I do not know of a good way to more concisely represent this data. There are four tasks with 3 models per task at a variety of distractor amounts. It seems like they need separate figures.} 

% From this, we see LeWM fail under the lowest disturbance amount tested, as LeWM has nothing to prevent the latent from collapsing under the irrelevant visual distractor. Across the other tasks, outside of OGBench Cube, we see a consistent pattern of SMWM holding up fairly well under increased distractor amounts until eventually collapsing, where our CF-JEPA holds up much better with minor performance degradation even at very large amounts of disturbance.
\added[id=SH]{LeWM already fails at the smallest number of distractors, since nothing in its objective prevents the latent from collapsing under task-irrelevant visual signals. SMWM is more resilient, but its performance drops sharply as the distractors become dense. On PushT, its success rate falls from 0.86 at $N\!=\!2$ to 0.27 at $N\!=\!64$, while CF-JEPA more gradually degrades from 0.87 to 0.57 in the same range. On TwoRoom, SMWM falls from 0.98 at $N\!=\!2$ to 0.46 at $N\!=\!32$, while CF-JEPA only falls from 0.95 to 0.82.} \added[id=MB]{The high variances \added[id=SH]{of SMWM} shown here indicate a mixture of degrading performance akin to the CF-JEPA behavior on some seeds and complete latent collapse on others. CF-JEPA \added[id=SH]{did not collapse in any of the configurations we tested}.}

\added[id=SH]{Interestingly, OGBench Cube is the only task that shows a different trend, where SMWM does not collapse even under heavy visual disturbance and retains 0.73 at $N\!=\!128$.} For OGBench Cube, we tested up to distractor amounts of $N\!=\!128$ before the exterior portion of the screen was entirely saturated with distractions and it still performed well. Even increasing the rectangle length, somewhat occluding the main robot and cube, still did not lead to SMWM collapse. We hypothesize that the specific task is robust due to its action space. Since it uses a differential IK solver for direct end effector control rather than something more low level like position control, all that is necessary for completing the control task is the goal latent.

\added[id=SH]{In the extreme case, the control success remains unchanged even when the scene is fully occluded or the camera is moved so that the robot is out of frame. This is because the pixel observations are not used for control.} Thus, the distractors have no effect on control performance unless there is encoder collapse to the point that the goal latent is meaningless, as in LeWM. \added[id=SH]{Since} this does not happen in SMWM or CF-JEPA, we see no failure.

\begin{figure}[t]
\begin{center}
\includegraphics[width=0.8\linewidth]{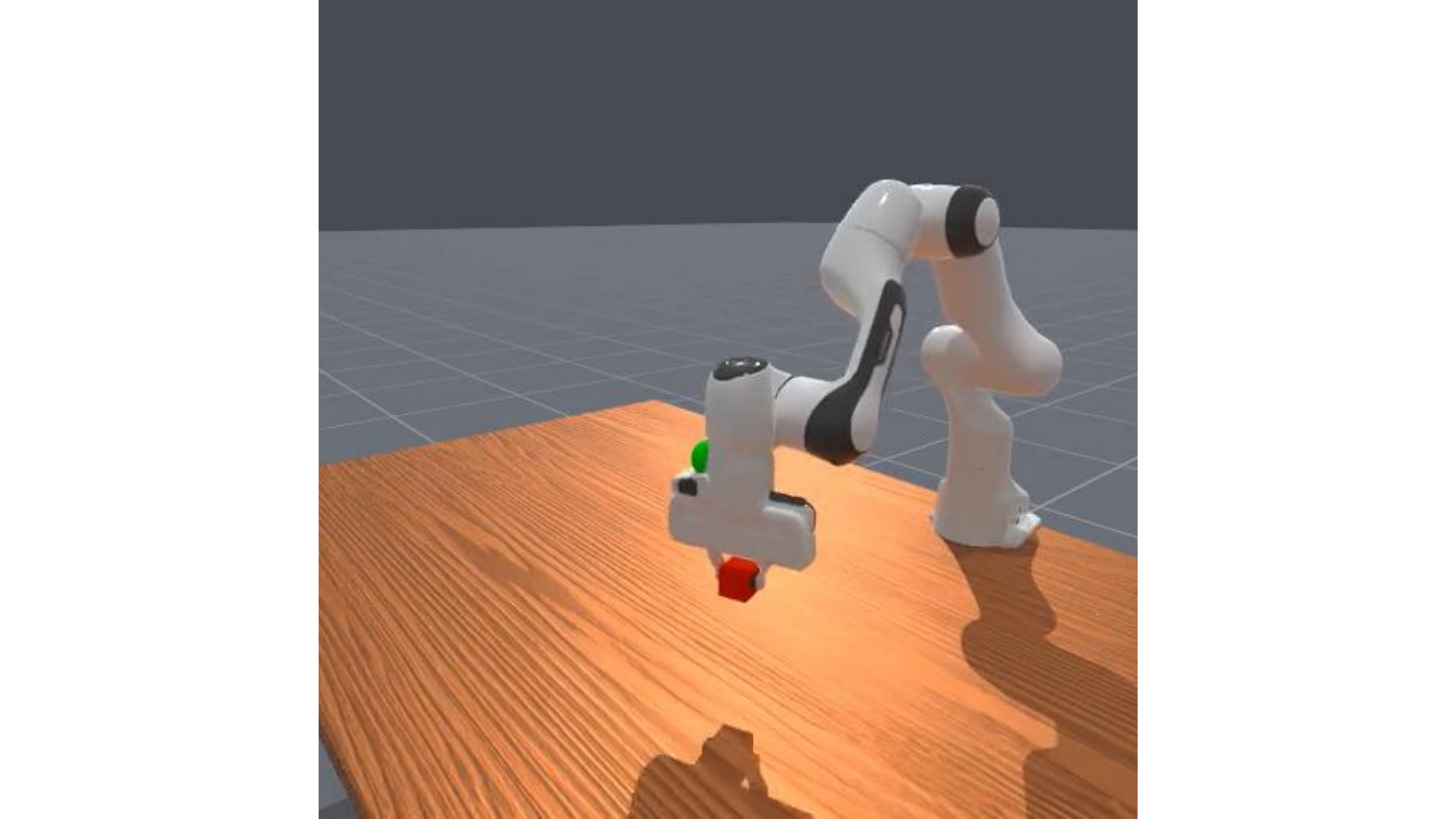}
\end{center}
\caption{Simplified ManiSkill Reach task used for further distractor evaluation.}
\label{fig:maniskill-reach}
\vspace{-1em}
\end{figure}

\input{tables/maniskill-distractor}
% Decoder reconstructions on Reacher as one grid: the target row once, then one output row per latent.
% Each source PDF (960x540 bp) holds a 2x3 grid; the trims select its top row (target) or bottom row (output).
\begin{figure}[t]
\centering
\setlength{\tabcolsep}{0pt}
\begin{tabular}{r@{\hspace{4pt}}c}
\scriptsize Target & \raisebox{-0.5\height}{\includegraphics[width=0.76\columnwidth, trim=60 275 64 1, clip]{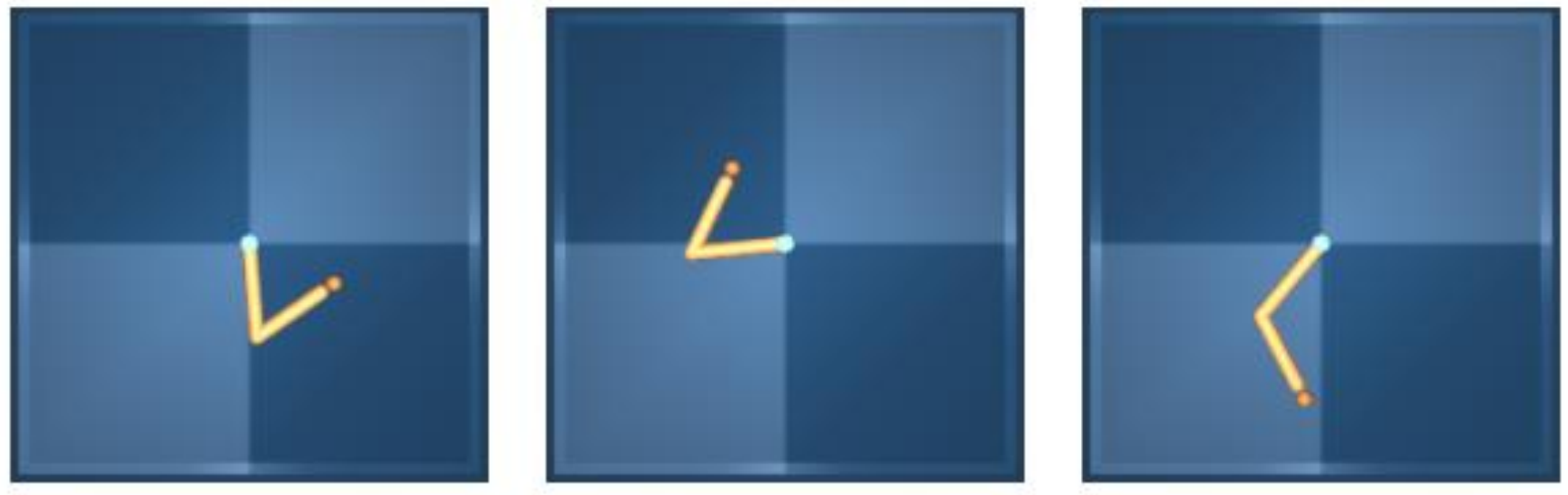}} \\[1pt]
\scriptsize CF-JEPA $z^c$ & \raisebox{-0.5\height}{\includegraphics[width=0.76\columnwidth, trim=60 4 64 265, clip]{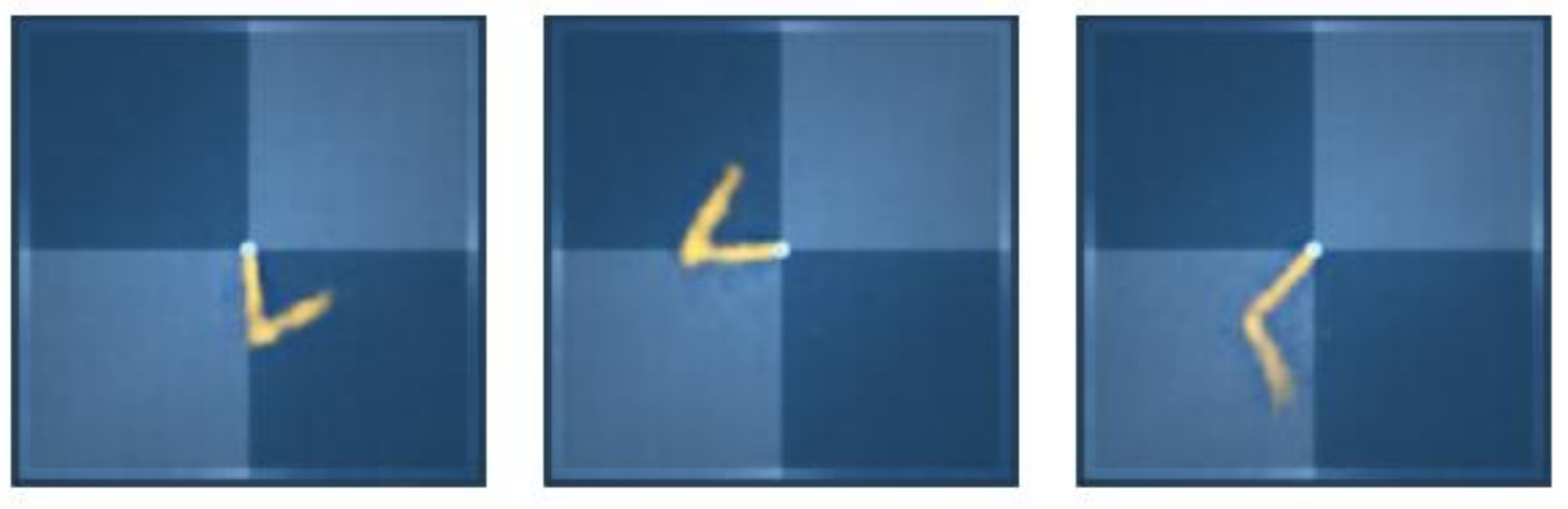}} \\[1pt]
\scriptsize CF-JEPA $z^u$ & \raisebox{-0.5\height}{\includegraphics[width=0.76\columnwidth, trim=60 4 64 265, clip]{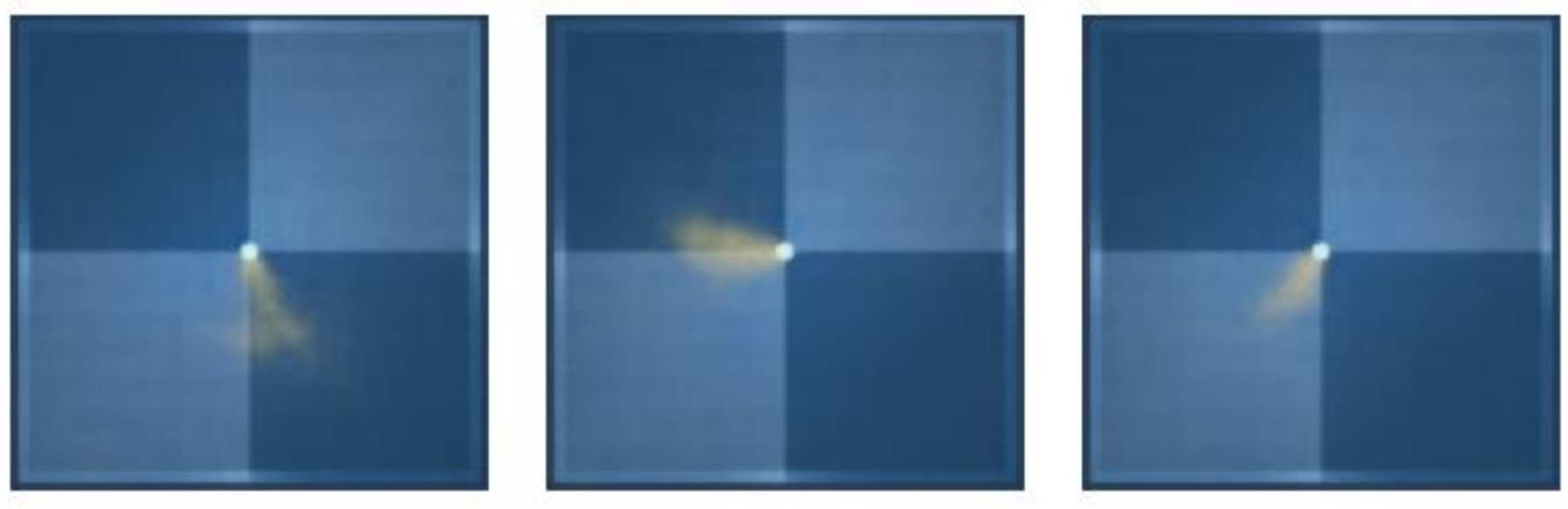}} \\[1pt]
\scriptsize SMWM & \raisebox{-0.5\height}{\includegraphics[width=0.76\columnwidth, trim=60 4 64 265, clip]{images/SMWM-Reacher-Reconstruction.pdf}} \\[1pt]
\scriptsize LeWM & \raisebox{-0.5\height}{\includegraphics[width=0.76\columnwidth, trim=60 4 64 265, clip]{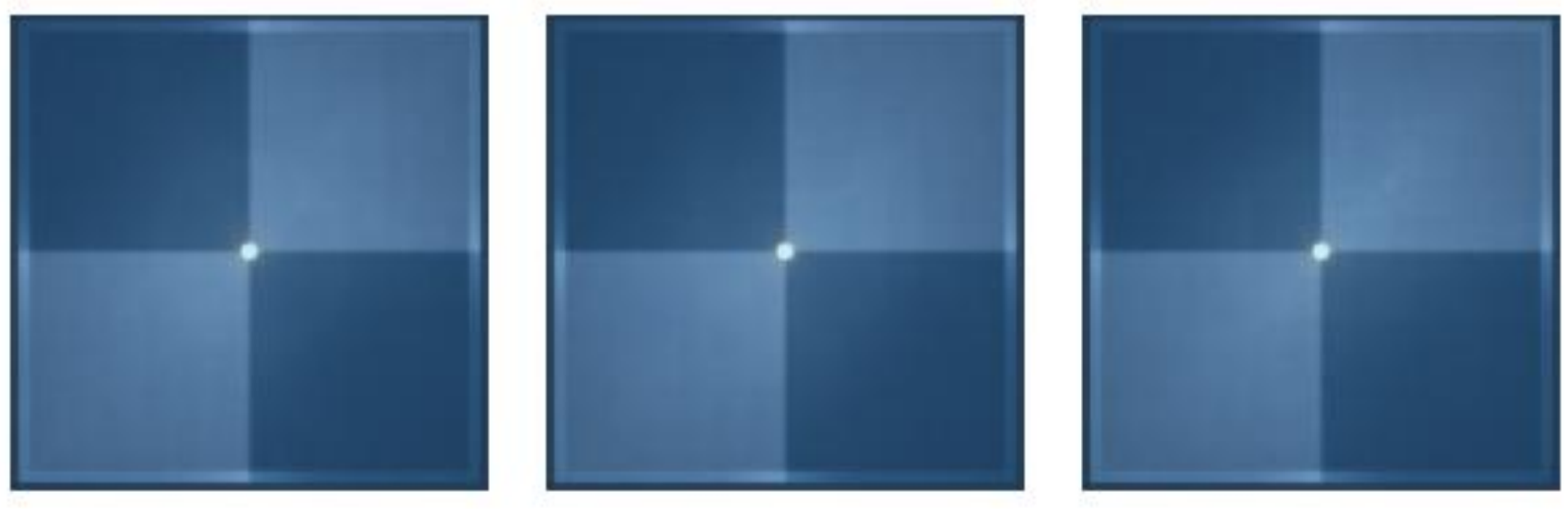}} \\
\end{tabular}
\caption{\protect\added[id=SH]{Only the controllable latent of CF-JEPA retains the agent. Reconstructions are decoded from frozen encoders trained with distractors and shown on Reacher test frames without distractors. Compared with the target (1st row), the controllable latent $z^c$ of CF-JEPA reconstructs the arm (2nd row), whereas the uncontrollable latent $z^u$ only blurs it (3rd row). SMWM and LeWM lose the arm entirely (4th and 5th rows).}}
\label{fig:reacher-decoder}
\vspace{-1.75em}
\end{figure}

\vspace{-0.5em}

\subsection{\protect\added[id=SH]{Performance on Perception-Dependent Tasks}}
\label{ssec:maniskill}
% Given that OGBench Cube was a poor task for demonstrating perception robustness, we proceeded with a targeted task in a simulated robotic domain to demonstrate viability of our method more broadly. We use a simplified variant of the ManiSkill~\cite{taomaniskill3} PickCube task, shown in Fig.~\ref{fig:maniskill-reach}, where we merely move the already gripped cube from its initial position to a goal position, similar to a reaching task. Given this, we collected a reduced dataset of 250,000 frames for training our models to highlight a failure of the baselines against ours. The success rates of our models under distractor conditions (N=16) are shown in Table~\ref{table:maniskill-distractor}. We show the success rates across a range of thresholds for reach accuracy, and our model outperforms until the threshold is so wide that both models have near perfect success.
\added[id=SH]{Because the action space of OGBench Cube bypasses the current observation, it cannot reveal whether a model perceives the agent robustly. We therefore evaluate on a task whose action space requires perceiving the state of the agent. We use a simplified variant of the ManiSkill~\cite{taomaniskill3} PickCube task (Fig.~\ref{fig:maniskill-reach}), in which the robot moves an already grasped cube to a goal position. Unlike OGBench Cube, the task uses joint position control, so the agent must infer the arm configuration from the image.} \added[id=SH]{We train all models on 250,000 frames and evaluate them with $N\!=\!16$ distractors. Table~\ref{table:maniskill-distractor} reports the success rate for different thresholds on the distance to the goal. At a loose threshold of 0.06~m, CF-JEPA and SMWM perform similarly, with a gap of only 0.04. As the threshold tightens, the gap widens to 0.15 at 0.03~m and 0.13 at 0.02~m. SMWM, which was as robust as CF-JEPA on OGBench Cube, falls behind once the task demands more precise perception.}

\vspace{-0.5em}

\subsection{Latent Space Analysis}
% Original paragraph (before merging the decoder figures into one grid), kept so it can be restored:
% We also performed analysis of the latent space generated by our model in comparison to the baselines. We trained a separate decoder network that takes as input the latent vector output from the frozen encoder network from each model with the goal of reconstructing the pixels from the original image. For a qualitative analysis, we look at the Reacher task. When incorporating distractors, Figs.~\ref{fig:reacher-decoder-zc} and \ref{fig:reacher-decoder-zs} show how effectively the control factorization works in a qualitative sense. It is clear from Fig.~\ref{fig:reacher-decoder-zc} that the arm is well reconstructed, ignoring the extraneous distractor information to focus on the important information for control. On the other hand, Fig.~\ref{fig:reacher-decoder-zs} is incapable of reconstructing the Reacher arm well, as this latent subspace cannot represent the controllable information. Looking at the baselines, Figs.~\ref{fig:smwm-reacher-decoder} and \ref{fig:lewm-reacher-decoder} show that under distractor conditions, LeWM and SMWM is collapsed and is entirely incapable of representing the Reacher agent. 
We also performed analysis of the latent space generated by our model in comparison to the baselines. We trained a separate decoder network that takes as input the latent vector output from the frozen encoder network from each model with the goal of reconstructing the pixels from the original image. For a qualitative analysis, we look at the Reacher task. \added[id=SH]{All models are trained with distractors, but we visualize the reconstructions on test frames without distractors.} \added[id=SH]{CF-JEPA learns to distinguish controllable information from uncontrollable information, as the 2nd and 3rd rows of Fig.~\ref{fig:reacher-decoder} show qualitatively.} It is clear from \added[id=SH]{the 2nd row} that the arm is well reconstructed, ignoring the extraneous distractor information to focus on the important information for control. On the other hand, \added[id=SH]{CF-JEPA with $z^u$ (the 3rd row)} is incapable of reconstructing the Reacher arm well, as this latent subspace cannot represent the controllable information. Looking at the baselines, \added[id=SH]{the 4th and 5th rows show} that under distractor conditions, LeWM and SMWM \added[id=SH]{are} collapsed and entirely incapable of representing the Reacher agent. 

As a quantitative measure consistent with the qualitative images in Figure~\ref{fig:reacher-decoder}, we use the participation ratio~\cite{Gao214262}, defined as:

\vspace{-0.5em}

\begin{equation}
\mathrm{PR}
=
\frac{\left(\sum_{i=1}^{d} \lambda_i\right)^2}
{\sum_{i=1}^{d} \lambda_i^2},
\label{eq:participation_ratio}
\end{equation}

where the eigenvalues $\lambda$ are computed from latents of a random 256 frames collected from the entire dataset, to demonstrate the spread of variance in the latent space for the clean and distracted settings. Table~\ref{table:latent-metric} shows this metric for each model on the Reacher task. From this, we can see that the distractors cause a large decrease for all models, with total collapse for SMWM and large decreases for LeWM and CF-JEPA. SMWM starts at a lower number as it has no SIGReg to spread out the latent space and falls to the minimum value of 1 due to collapse. Neither LeWM or CF-JEPA fall to zero due to SIGReg, but since LeWM falls 21.5 times compared to the 3.3 times for CF-JEPA, it is not performant for control or decoding, leaving CF-JEPA the only functional model tested.

\input{tables/latent-metric}

\vspace{-0.5em}

%% file: tables/task-with-distractors.tex
\begin{table}[]
\vspace{0.3em}
\centering
\caption{Task performance under standard distractor settings (N=2) For all tasks, we report mean $\pm$ one standard deviation.}
\label{table:task-with-distractor-performance}
\begin{tabular}{l|l|l|l}
\hline
Task & LeWM & SMWM & CF-JEPA \\ \hline
Reacher & \meanstd{0.07}{0.05} & \meanstd{0.1}{0.03} & \meanstd{0.48}{0.12} \\ \hline
PushT & \meanstd{0.01}{0.01} & \meanstd{0.86}{0.02} & \meanstd{0.87}{0.03} \\ \hline
OGBench Cube & \meanstd{0.42}{0.02} & \meanstd{0.84}{0.01} & \meanstd{0.85}{0.02} \\ \hline
TwoRoom & \meanstd{0.33}{0.03} & \meanstd{0.98}{0.05} & \meanstd{0.95}{0.03} \\ \hline
\end{tabular}
\vspace{-0.5em}
\end{table}

%% file: tables/task-with-distractor-sweep.tex
\begin{table}[]
\vspace{0.3em}
\centering
\caption{\protect\added[id=SH]{Task performance as the number of distractors $N$ increases. For each task, we report values up to the collapse of SMWM, except for OGBench Cube, where SMWM does not collapse.}}
\label{table:distractor-sweep}
\begin{tabular}{l|l|l|l}
\hline
Task & LeWM & SMWM & CF-JEPA \\ \hline\hline
Reacher ($N\!=\!2$)   & 0.08 & \meanstd{0.1}{0.03} & \bmeanstd{0.48}{0.12} \\ \hline\hline
% Reacher ($N\!=\!8$)   & -- & -- & -- \\ \hline
% Reacher ($N\!=\!16$)  & -- & -- & -- \\ \hline
% Reacher ($N\!=\!32$)  & -- & -- & -- \\ \hline
% Reacher ($N\!=\!64$)  & -- & -- & -- \\ \hline
% Reacher ($N\!=\!128$) & -- & -- & -- \\ \hline\hline
PushT ($N\!=\!2$)   & 0.01 & \meanstd{0.86}{0.02} & \bmeanstd{0.87}{0.03} \\ \hline
% PushT ($N\!=\!8$)   & -- & -- & -- \\ \hline
PushT ($N\!=\!16$)  & -- & \meanstd{0.36}{0.36} & \bmeanstd{0.76}{0.} \\ \hline
PushT ($N\!=\!32$)  & -- & \meanstd{0.57}{0.01} & \bmeanstd{0.68}{0.06} \\ \hline
PushT ($N\!=\!64$)  & -- & \meanstd{0.27}{0.25} & \bmeanstd{0.57}{0.01} \\ \hline\hline
% PushT ($N\!=\!128$) & -- & -- & -- \\ \hline\hline
OGBench Cube ($N\!=\!2$)   & 0.41 & \meanstd{0.84}{0.01} & \bmeanstd{0.85}{0.02} \\ \hline
% OGBench Cube ($N\!=\!8$)   & -- & -- & -- \\ \hline
% OGBench Cube ($N\!=\!16$)  & -- & -- & -- \\ \hline
OGBench Cube ($N\!=\!32$)  & -- & \meanstd{0.80}{0.02} & \bmeanstd{0.81}{0.01} \\ \hline
OGBench Cube ($N\!=\!64$)  & -- & \meanstd{0.70}{0.05} & \bmeanstd{0.73}{0.01} \\ \hline
OGBench Cube ($N\!=\!128$) & -- & \bmeanstd{0.73}{0.05} & \meanstd{0.70}{0.02} \\ \hline\hline
TwoRoom ($N\!=\!2$)   & \meanstd{0.34}{0.05} & \bmeanstd{0.98}{0.05} & \meanstd{0.95}{0.03} \\ \hline
TwoRoom ($N\!=\!8$)   & -- & \bmeanstd{1}{0} & \meanstd{0.97}{0.01} \\ \hline
TwoRoom ($N\!=\!16$)  & -- & \meanstd{0.6}{0.28} & \bmeanstd{0.94}{0} \\ \hline
TwoRoom ($N\!=\!32$)  & -- & \meanstd{0.46}{0.12} & \bmeanstd{0.82}{0} \\ \hline
% TwoRoom ($N=64$)  & -- & -- & -- \\ \hline
% TwoRoom ($N=128$) & -- & -- & -- \\ \hline
\end{tabular}
\vspace{-0.5em}
\end{table}

%% file: tables/maniskill-distractor.tex
\begin{table}[]
\centering
\caption{Task performance under N=16 distractor settings for ManiSkill Reach task. We report mean $\pm$ standard deviation across three seeds for each model.}
\label{table:maniskill-distractor}
\begin{tabular}{l|l|l|l}
\hline
Success Threshold (m) & LeWM & SMWM & CF-JEPA \\ \hline
0.06 \added[id=SH]{(easiest)} & \meanstd{0.55}{0.04} & \meanstd{0.79}{0.05} & \bmeanstd{0.83}{0.06} \\ \hline
0.05 & \meanstd{0.51}{0.05} & \meanstd{0.62}{0.04} & \bmeanstd{0.72}{0.08} \\ \hline 
0.04 & \meanstd{0.44}{0.03} & \meanstd{0.58}{0.02} & \bmeanstd{0.64}{0.1} \\ \hline
0.03 & \meanstd{0.36}{0.03} & \meanstd{0.38}{0.02} & \bmeanstd{0.53}{0.14} \\ \hline
0.025 & \meanstd{0.32}{0.06} & \meanstd{0.30}{0.04} & \bmeanstd{0.48}{0.17} \\ \hline
0.02 \added[id=SH]{(hardest)} & \meanstd{0.21}{0.05} & \meanstd{0.22}{0.04} & \bmeanstd{0.35}{0.12} \\ \hline
\end{tabular}

% \protect\sehoon{Let's report only thresholds up to 0.06 and reverse the row order, from wide to narrow. The story is that the gap is only 0.04 at 0.06 but grows as the threshold narrows to 0.03 or 0.02.}}
\vspace{-1em}
\end{table}

%% file: tables/latent-metric.tex
\begin{table}[]
\centering
\caption{Latent vector participation ratio under nominal and distracted conditions for the Reacher task. LeWM and SMWM both collapse much further than CF-JEPA, correlating with the worse decoder and control performance.}
\label{table:latent-metric}
\begin{tabular}{l|l|l|l}
\hline
 & LeWM & SMWM & CF-JEPA \\ \hline
Nominal & \meanstd{48.2}{0.8} & \meanstd{3.71}{0.07} & \meanstd{42.3}{0.1} \\ \hline
Distractor (N=2)  & \meanstd{2.24}{0.23} & \meanstd{1.}{0.01} & \meanstd{12.9}{0.35} \\ \hline
\end{tabular}

% \protect\sehoon{Let's report only thresholds up to 0.06 and reverse the row order, from wide to narrow. The story is that the gap is only 0.04 at 0.06 but grows as the threshold narrows to 0.03 or 0.02.}}
\vspace{-1.25em}
\end{table}

%% file: section/conclusion.tex
\section{CONCLUSION}

This paper presents a method for training a JEPA world model with improved robustness to visual disturbances. We build our CF-JEPA off of LeWM, incorporating additional structure in the latent space via adding inverse dynamics losses so that we can factorize the space into controllable, action-conditioned, and uncontrollable, action-independent, spaces. We train our model for a set of four tasks under nominal conditions and under conditions with task-irrelevant distractor signals. Our model shows comparable performance against the baselines under nominal conditions, and improved performance under distracted conditions, being the only model tested that did not experience latent collapse.

\added[id=SH]{Our experiments suggest two directions for future work. First, we validate CF-JEPA in simulation with a single family of distractors, and applying it to a real robot would test the factorization under richer visual conditions. Second, we assume that the uncontrollable part of the scene never affects the controllable part, so settings where the two interact, such as a person working next to the robot, would require extending the model.}